\documentclass[12pt]{article}
\usepackage{hyperref}
\usepackage[a4paper, lmargin=1.0in, rmargin=1.0in, tmargin=1.0in, bmargin=1.0in]{geometry}
\begin{document}


\title{Thoughts on an Unified Framework for Artificial Chemistries}
\author{{\bf Janardan Misra}~\thanks{Work done when author was in NUS (2002-2005).} \\ HTS (Honeywell Technology Solutions) Research Lab\\ 151/1 Doraisanipalya, Bannerghatta Road,\\ Bangalore 560 076, India\\
{\sf Email: janardan.misra@honeywell.com}
}

\date{}
\maketitle
\tableofcontents

\begin{abstract}
Artificial Chemistries (ACs) are symbolic chemical metaphors for the exploration of Artificial Life, with specific focus on the problem of {\it biogenesis} or the origin of life. This paper presents authors thoughts towards defining a unified framework to characterize and classify symbolic artificial chemistries by devising appropriate formalism to capture semantic and organizational information. We identify three basic high level abstractions in initial proposal for this framework viz., \emph{information, computation, and communication}. We present an analysis of two important notions of information, namely, Shannon's Entropy and Algorithmic Information, and discuss inductive and deductive approaches for defining the framework. 
\end{abstract}

\section{Basic Framework of Artificial Chemistries}

Aim of this section is to present a brief introduction to
artificial chemistries. We will start with a discussion on the epistemological
foundations of the area and will illustrate further details using examples
relevant to this proposal. The examples are followed by discussions
to motivate the main theme of the proposal which is elaborated in
coming sections.

\subsection{Introduction}

It is a long held topic of scientific debate whether there are any
biological principles of life and other complex biological
phenomena, which are not directly reducible to physical and
chemical laws. Living beings, however small and consisting of the
same molecular components as non living things, nonetheless
exhibit qualitatively different characteristics. This may be in
part due to the complex organizational structure which
distinguishes them or it could be because of their quantitatively
complex structure which gives rise to difficulty in analyzing
properties using currently available tools.

The direct ways to understand this complex biological phenomena
are usually difficult and error prone because living structures
are by default complex and hard to manipulate. Even cellular level
experiments are difficult to carry out and their simulations are
quite cumbersome.

\emph{Artificial life} (AL) is a tool to study principles
explaining this complex phenomena of life without directly getting
involved with the real biological systems. The fundamental
assumption here is that principles of life are independent of the
medium and carbon based life on earth is just one example of the
possible forms of life. This means even artificial environments
like digital media can also exhibit life-like behaviour under
certain conditions. This way AL complements the main stream
biological studies by synthesizing life-like systems using digital
media. There are several such examples where these artificial life
forms exhibit properties remarkably close to higher forms of life,
e.g., Tierra \cite{tierra}, Avida \cite{1}.

Living phenomena has several aspects to study, one such is the
origin of life or \emph{biogenesis}. Here the problem is to
understand how first primitive form of life such as metabolism and
self replicating structures could have come into existence
starting from non living chemical compounds. \emph{Artificial
chemistries} (AC) are the primary tools in AL studies aimed at
understanding this origin of life and other complex emergent
phenomena. ACs follow chemical metaphor. Like real chemical
reactions between molecules, which give rise to new molecules, ACs
as well define abstract molecules and reactions and study what
emerges during the course of reactions.

An AC has three main components, a set of objects or
\emph{molecules}, a set of \emph{reaction rules} or collision
rules, and a definition of \emph{population dynamics}.

Objects can be abstract symbols, numbers, lambda expressions,
binary strings, character sequences, abstract data
structures etc. Reaction rules might be string matching,
string concatenation, reduction rules, abstract finite state
machines, Turing machines, matrix multiplication, simple
arithmetic operation, cellular automata, boolean networks etc.
Dynamics can be specified in terms of ordinary differential
equation, difference equation, meta dynamics, explicit collision
simulation, well stirred reactor, self organizing topology, etc.

A survey on various ACs is given in \cite{12}, which also has some
broad classification of ACs based upon the kind of molecular
abstractions (explicit or implicit), type of reaction rules
(constructive or non constructive), and population dynamics.

To illustrate, we take examples from two kinds of ACs. One where
no spatial structures are considered, that is, all molecules
evolve as a whole in a reactor tube and all molecules can interact
with each other according to the collision rules. The examples of
AlChemy (Section~\ref{alchemy}) and CHAM/ARMS (Section~\ref{cham}) are of this
type. Second kind of AC systems impose some sort of spatial
structures on the molecules thus limiting the possible reactions
between molecules to their ``neighbourhood'' only. Planar graph
(Section~\ref{pgraph}) based AC is of this type.

It seems, during the pre-biotic evolution of life, spatial
structures (e.g., membranes etc) emerged starting from the open
reactor type system without any spatiality. This spatial structure
based classification is one of the main motivations for P system
based AC definition, we propose in the next section.

\subsection{Examples}

Next we illustrate the common design of ACs using examples. Each example is followed by a discussion on the relative strengths and limitations of it w.r.t. real chemistry. 

\subsubsection{Algorithmic Chemistry - AlChemy}~\label{alchemy}
We consider $\lambda$ expression based AC proposed in \cite{3,4}
called AlChemy.\\\\
\textbf{Molecules - $\lambda$ Terms:} The object space consists of abstract lambda expressions
 (also called \emph{terms}). These terms are generated as follows: There is an infinite supply of variable names $V$ = $\{x, y, z,\ldots\}$. Other than $V$, the alphabet consists of a lambda symbol '$\lambda$', dot '$.$', and encapsulating brackets '$($', '$)$'.

  The set of terms, $\Lambda$, is defined inductively:

(1) $x \in V \Rightarrow x \in \Lambda$

(2) $x \in V; M \in \Lambda \Rightarrow \lambda x.M \in \Lambda$
(abstraction)

(3) $M \in \Lambda; N \in \Lambda \Rightarrow (M)N \in \Lambda$
(application)

A variable $x$ is said to be bound if it occurs inside a sub-term
with the form $\lambda x.P$, otherwise it is free. The set of free
variables in an expression $P$ is denoted by $f(P)$.

\emph{Syntactical Transformation: }The schemes of transformation
are oriented rewrite rules. Structures on the left ­ hand side are
replaced by structures on the right ­ hand side. More precisely,

\emph{Substitution}

(4) $(\lambda x.x)Q \rightarrow Q$

(5) $(\lambda x.E)Q \rightarrow E$; if $x \not\in f(E)$

(6) $(\lambda x.\lambda y.E)Q \rightarrow \lambda y.(\lambda
x.E)Q$; if $x \neq y$ and $(x \not\in f(E) \vee y \not\in f(Q))$

(7) $(\lambda x.(E_1 )E_2 )Q \rightarrow ((\lambda x.E_1
)Q)(\lambda x.E_2 )Q$

\emph{Renaming}

(8) $\lambda x.E \rightarrow \lambda z.(\lambda x.E)z; z \not\in
f(E)$\\\\
\textbf{Reaction Rules - Function Composition and Normal Form
Reduction:} The reaction rules in Alchemy consist of application of one lambda
term over the other, which is then reduced to a normal form. The
choice of lambda calculus allows the abstract formulation of
chemical substitution during chemical reactions. Normalization is
used to get equivalence classes based on functional equivalence.
Since normal form reduction is undecidable in case of lambda
calculus, reduction steps are finitely bounded \cite{4}.

Formally a reaction between molecules $A$ and $B$ can be written
as a binary operation ($+_\Phi$) defined as

$$A +_\Phi B \rightarrow A + B + nf(((\Phi)A)B)$$

Where $+$ is used from the convention of writing chemical
equations to represent that the molecules are present in reactor.
$nf()$ uses some consistent reduction strategy to reduce the term
in finitely many steps to a normal form. This choice of finite
step normal form reduction actually results in equivalence classes
consisting of all the expressions which are functionally same
modulo finite execution steps. The choice of $\Phi$ gives
flexibility in the way molecules can react.\\\\
\textbf{Population Dynamics - Stochastic Molecular Collisions:}
Initially a large pool of random lambda terms of finite lengths is
generated. Only those terms, which are in normal form are
considered. In each iteration two molecules are chosen at random
and one is applied to the other (function composition) according
to $\Phi$, which is fixed at the beginning. Result is reduced to
its normal form in finite steps. Filtering conditions are applied,
for example, before collision takes place if the operator molecule
does not start with symbol $'\lambda'$ then it is discarded. These
filter conditions are basically meant to ensure consistency in
results as per the lambda calculus semantics and to give diversity
to the emerging organizational structures. Flow is maintained by
randomly selecting molecules and removing them from the reactor.

The relative quantitative dynamics of various molecules is
captured in terms of differential equations. Replicator equations
of Lotka-Volterra type \cite{3, 5} are used to describe the
relative concentration of self replicating molecules. \\\\
\textbf{Discussion:} In actual chemistry, especially in case of organic compounds with
 chains of carbon atoms and possible branching, chemical reactions
 substitute parts of one molecule with other molecule thus leading
 to structural rearrangement in the chemical composition of these
 molecules. This is the main motivation behind the choice of
 lambda terms in AlChemy, where the substitution is abstracted as function composition of lambda terms. Second motivation is that many chemical reactions can give rise to the same chemical compound, which is captured by normal form equivalence. The AlChemy is also a constructive chemistry like real chemistry.  Also the notion of equality leads to formation of network of molecules.

 With stochastic collision dynamics and choice of reaction type ($\Phi$), the AlChemy gives rise to some interesting forms of organizations, classified as \emph{level-0}, \emph{level-1}, and \emph{level-2} organizations. While \emph{level-0} organization consists of only self replicating molecules whose frequencies are modeled using replicator equations, \emph{level-1} organization has strong element of self-maintenance where any reaction between two molecules produces a new molecule inside the
 same population. \emph{level-2} organization is a coexistence of two interdependent \emph{level-1} organizations which support each other.

 Though AlChemy captures certain basic aspects of real chemical
 compounds and their reactions, it has its own limitations.
 Most important of those is related to the choice of lambda calculus.
 Even though lambda calculus is computationally universal and has a
 consistent reduction  strategy (i.e., order of reduction steps does not change the result),
 it has no serious bearing on its chemical counter part. Actual chemical reactions are not only much more complex,
 they might not follow computationally consistent mechanisms like total substitution.

 Thus the first limitation is the lack of
 selective substitution, which means, in case of actual
 chemical reactions, new compounds are formed (with substitution)
 based on the relative strengths of chemical bonds in reactants
 and relatively higher stability of the products. On the other hand in case of
 substitution in lambda terms no such conditions apply and instances
 of free variable are equally substituted everywhere.
 We propose alternate structure and reaction rules to overcome
 this limitation in the next section.

Second limitation is the poor abstraction of structural
    properties of chemical compounds. The only kind of compounds
    which might be resembling the lambda terms structurally are
    those which have long carbon chains with possible branching.
    Double helix structure of DNA
    with complementarity is difficult to capture using
    lambda terms. Other geometrical properties like chirality
    \footnote{Many important molecules required for life exist in two forms.
    These two forms are non-super imposable mirror images of each other,
    like the left and the right hand. This property is called
    chirality.}, which is so common in living
    forms \footnote{Nearly for all biological polymers to function
    their component monomers must have the same handedness.
    All amino acids in proteins are 'left-handed',
    while all sugars in DNA and RNA, and in the metabolic
    pathways, are 'right-handed'},
    as well cannot be captured using lambda terms. The significance of
    this lack of structural abstraction of geometrical properties is
    not very clear.

    Since chemical reactions are driven by thermodynamic
    constraints like rate of collision, pressure etc, and the properties of colliding
     molecules, they are usually symmetrical in nature. Thus the result of
     collision of the molecules A and B
     is same as that of B and A since there is no order
     on A and B.
      On the other hand, that is not the case with function composition,
    which is in general asymmetric in its definition.
    In our view, this presence of asymmetry
    in lambda chemistry might detach it from the real chemistry significantly.

    Functional Equivalence - the kind of functional
    equivalence defined in case of lambda chemistry does not
    capture the equivalence which life-like forms
    demonstrate. In case of living structures, it is the
    interaction which objects have with external environment or other
    objects that plays important role. This element of interaction is
    not captured well. One idea is to consider $\pi$ - calculus like
    formalism \cite{Parrow01} which has $bisimulation$ kind of
    equivalence which can be used to capture the equivalence in the objects based
    upon how they  can interact with other objects.

    Lack of information abstraction - this is true in
    general for almost all of the proposed ACs. And that is
    one of the focus of this proposal to understand the role
     information plays in the emergence of life-like phenomena in ACs.

\subsubsection{The Chemical Abstract Machine}~\label{cham}

The Chemical Abstract Machine (CHAM) was proposed in \cite{cham96}
as an abstract formalism for concurrent computation using closely
a metaphor of chemical reactions.

There are two description levels. On the upper level, CHAM
abstractly defines a syntactic framework and a simple set of
structural behavior laws. An actual machine is defined by adding a
specific syntax for molecule and a set of transformation rules
that specify how to produce new molecules from old ones. \\\\
\textbf{Molecules} are \emph{terms of some algebra}. A general
membrane construct transforms a solution into a single molecule,
and an associated general \emph{airlock} construct makes the
membrane somewhat porous to permit communication between an
encapsulated solution and its environment. The \emph{generic
reactions laws} specify how reactions defined by specific
transformation rules can take place and how membranes and airlocks
behave. A specific machine is defined by giving the algebra of
these terms and the rules. Not all molecules directly exhibit
interaction capabilities. Those which do are called \emph{ions}.
The interactive capability of an ion is generally determined only
by a part of it that is called its \emph{valence}.\emph{ The
reaction rules} are used to build new molecules from the ions. The
non-ion molecules can be heated as per the \emph{heating rules} to
break them into simpler sub-molecules. Conversely, a set of
molecules can cool down to a complex molecules using reverse
\emph{cooling rules}. The presence of membrane type structure
gives universal computational power to the model. \emph{Dynamics}
of CHAM goes like this - on each iteration a CHAM may perform an
arbitrary number of transformations in parallel, provided that no
molecule is used more than once to match the left side of a
reaction law. A CHAM is non-deterministic if more than one
transformation rules may be applied to the population at a time.

Sujuki and Tanaka used CHAM to model chemical systems by defining
an ordered abstract rewriting system on multiset called chemical
ARMS \cite{arms}. \emph{Molecules} are the \emph{abstract
symbols}. The \emph{reaction rules} are \emph{multiset rewriting
rules}. The reactor is represented by a multiset of symbols with a
set of input strings. An optional order is imposed on the rules,
which specifies in which order the rules are processed. Different
rate constants are modelled by different frequencies of rule
application. \\\\
The qualitative \textbf{dynamics} of ARMS is investigated by
generating rewriting rules randomly.  This led them to derive a
formal criteria for the emergence of cycles \cite{armscycles} in
terms of an order parameter, which is roughly the relation of the
number of heating rules to the number of cooling rules
\cite{armsorder}. For small and large values of this order
parameter, the dynamics remains simple, i.e., the rewriting system
terminates and no cycles appear. For intermediate values, cycles
emerge.\\\\
\textbf{Discussion:} Although CHAM was not defined as an AC, it is quite close to
actual cellular chemistry in some aspects. The presence of
membrane structure gives rise to important resemblance with
cellular reactions mediated by membranes. Another significant
property of CHAM model is that it is very general hence provides
flexibility in the way actual model is defined. Heating and
cooling laws closely capture what happens in case of actual
chemical reactions under the effect of temperature.

The main limitation of CHAM model is that the allowed abstract
terms of algebra are not adequate to capture the structural
properties of real chemical compounds, as discussed in case of
AlChemy.

Second limitation comes due to nature of rewriting rules, they are
actually grammar rules rather than being close to the chemical
reactions. Because of this problem with multiset rewriting, in
ARMS analysis is done by randomly generating these rewriting laws,
and it is not clear whether chemical reactions where molecules
actually interact and forge new bonds or break up can be fully
modeled this way.

\subsubsection{Artificial Chemistry on a Planar Graph}\label{pgraph}

This model of AC was proposed in \cite{planargraph}, where an AC
is embedded in a planar triangular graph. Molecules are placed on
the vertices of the undirected graph and interact with each other
only via the edges. The planar triangular graph can be manipulated
by adding and deleting nodes with a minimal local rearrangement of
the edges. The graph based approach provides handle for spatial
structures.\\\\
\textbf{Molecules and Reactions:} There is an (infinite) set of
potential molecules $S$ and a reaction mechanism which computes
the reaction product for two colliding molecules $x, y \in S$.
There may be an arbitrary number of products for each such
collision. Molecules are built from different types of substrate
of elements called atoms. Each type is associated with a different
function. The total number of atoms in the reactor is kept
constant during a run. Free atoms (not bounded in molecules) are
separately stored and form a global pool. \\\\
\textbf{Dynamics:} At every step they pick two neighbouring molecules
$(x, y)$ and apply the first $x$ to the second $y$ creating a
(multi­)set of new molecules. These product molecules are randomly
inserted in the two faces next to the link between $x$ and $y$.
$x$ is replaced with first molecule after the reaction (the result
of the combinator reduction) and $y$ is finally deleted. Molecules
cannot change their positions in the graph.

In this system, it is observed that clusters of
molecules which do not interact with the neighbouring molecules
arise. The clusters can be regarded as membranes when they divide
the graph into different regions. There also arises a cell
organization, that is, a subgraph that can maintain the membranes
by themselves.\\\\
\textbf{Discussion:} As noted in \cite{12}, the presence of spatial topology gives rise
to certain phenomena which is not possible to emerge easily in
cases where there is no spatial topology present in the model. For
example in the case of this planar graph based AC, an emergence of
membrane type structure is something which is frequently observed
in living systems. This phenomena does not emerge in open reactor
type of ACs with no spatial structures. Another important property
is the emergence of self organization in the form of maintaining
the membrane structures. Choice of "atom - symbols" as basic
molecular unit closely resembles real chemical composition of
molecules consisting of atoms.

On the other hand, the choice of planar graph based topology is
not something usually present in cellular structures neither it
can be a simplified spatial structure for initial chemical
environment responsible for emergence of life. Absence of
abstraction of geometrical or structural properties is yet another problem.

\subsection{More Discussion on Artificial Chemistries}

ACs are basically motivated by and developed to understand the
pre-biotic evolution or the problem of origin  of life, which is
still an open problem despite lots of advancements in molecular
biology \cite{6, 7, 8}. The problem of pre-biotic evolution
differs significantly from the post-biotic phenomena mainly
because of the appearance of genetic material. Once the first form
of life, a single cell or more primitive forms are available,
Darwinian theory of evolution based upon mutation and selection
\cite{9} or neutral theory of random drifts \cite{10}, etc can be
used to explain the emergence of higher and more complex forms of
life. Still the emergence of this genetic material which is so
fundamental for the proper functioning of even the simplest forms
of life is what makes the problem of pre-biotic evolution so
different.

Therefore the kind of problems mainly of focus in ACs and in this
proposal are the search for principles governing the emergence of
life-like forms from non life-like structures in AC systems. This
also involves proper level of abstraction from real chemistry
without loosing generality. 

In AC, we primarily consider the qualitative aspects of a problem,
before considering the quantitative relations between its
components. The quantitative aspect is usually analyzed using
reactor flow equations \cite{11}.  The stable structures generated
by artificial chemistries, the stable sets of molecules, are
usually referred to as \emph{organizations}. Understanding which
organization will appear is one such example to understand the
qualitative solution of an AC.

Some of the aspects very commonly studied in AC are - given an AC,
how to know a priori, which organizations are possible and which
are not possible? To know which organizations are probable and
which are improbable? To define an AC to generate a particular
organization? How stable are organizations? Can the complexity of
an organization be defined? If is possible to generate an AC which
moves from organization to organization in a never ending growth
of complexity? Quantitative questions can also be asked, for
example, given an AC, in a particular organization how many stable
(attractive) states are present inside it?

\cite{12} has detailed description of several interesting common
phenomena which are observed in different kinds of AC systems such
as reduction of diversity, formation of densely coupled stabled
networks, syntactic and semantic closure in these networks etc.

\section{An Abstract Framework for Artificial Chemistries }

One of the pressing needs of AC studies is to develop an unified
framework to understand the role of various ACs from the point of
view of their basic aim of explaining the possible principles
leading to the origin of life-like structures. If we look into the
varying nature of molecular abstractions used in various ACs, the
varying definitions of reaction rules and population dynamics, we
find that it is difficult to understand why only certain phenomena
emerges in one AC set up but not the other, which might be
emerging with some other AC. There is no single AC which gives
rise to all important life-like properties. The role of spatial
structures is one such example, where we notice that only the presence
of initial topological constraints make it possible the emergence
of realistic cellular forms in emerging organizations.
  Thus the aim of this section is to motivate the need for a new unified
framework to characterize and classify symbolic artificial
chemistries with appropriate formalism. We explain three basic
high level abstract components identified in our initial proposal
for this unified framework.

\subsection{Do Artificial Chemistries Correctly Abstract the Real Chemistry?}

The problem of epistemological cut is deeply present in any branch
of AL \cite{14}. This is true in the case of AC as well. The
problem is up to what extent the definition of an AC should be
based on real chemistry to demonstrate that certain kind of life
like phenomena emerges even when molecules or objects are not
exactly the real chemical compounds.  Thus as we analyze, we
consider only those ACs, which are similar to some extent with the
real chemical systems, e.g., AlChemy, CHAM, etc. Even in case of
these ACs which aim to abstract closely the real chemical
environment, we find that they do not come sufficiently close to
assert claims in generic sense. In the spectrum of ACs there are
examples of those which demonstrate several of high level
organizational properties, for example origin of diversity of
life, in Tierra \cite{tierra}, but the power comes out of
\emph{in-built} self replicating and self organizing properties in
the basic structures (programs). On the other hand we have
examples which closely simulate the bio chemical reactions, e.g.,
self assembly of protocell structures, but these are complex, time
consuming, and do not explain the emergence of complex
organizational patterns or life-like properties. This motivates
for the need of correctly abstracting the most essential and basic
properties from real chemical environment and to explore dynamic
structures in an unified way.

 \subsection{Need and Structure of a Unified Framework}

 The most basic aspects of emergence of life from non living
 matter are the emergence of \emph{replicative mechanism } and the
 emergence of \emph{metabolism}. Replicative mechanism emerges as a tool for
 preserving the structure and function under the disintegrating effect
 of second law of thermodynamics. This replicative mechanism as we
 know it, in case of almost all forms of life, consists of two basic and
 functionally different components \cite{7,8} - one that encodes
 the instructions   how to replicate, and the other, which actually
 carries out the actual task. These are nucleic acids and proteins
 respectively. The emergence of \emph{metabolism} keeps the cells in
 thermodynamic equilibrium. The metabolism is important
 because otherwise a cell will disintegrate soon under external
 perturbations.

 Emergence of either the replication or the metabolism as autocatalytic reaction
 networks \cite{kuffman86} is demonstrated in case of most of the AC
 systems, along with other complex organizational forms mainly suitable to
 be compared with post biotic life forms \cite{12}. Even then the emerging organizations
 usually do not come very close to the existing cellular structure. The reason might be
 that either the molecular structure or the collision rules defined in
 these ACs do not match very well with the real chemistry, as discussed in case of AlChemy
 in the previous section .  In cases where real chemical systems are simulated, simulations
  become quite complex, time consuming and exact analysis of the results is not easy.

  At present there is no such standard analytical framework existing, which can be used to understand these aspect in a unified way. To take particular example, we consider the ``genetic'' information. \cite{11} cites the \emph{sequence hypothesis} to explain the qualitatively different role played by genetic material inside a cell. The sequence hypothesis states that generation and functioning of certain bio molecules is totally controlled by genetic material and thus, the functioning of cellular processes cannot be explained only in terms of chemical and physical properties of cell compounds. The point is that genetic sequence encodes some very specific instructions which actually direct those processes. Now once we know that this happens, we can try to understand these coded instructions or explain why this works that way using the physical and chemical laws but this does not explain why and how that structure emerged in the first place and why only that way. In AC parlance, this amounts to developing suitable framework which can explain the principles behind all this phenomena without working exactly with real chemistries. In this proposal we aim to address this problem  by formally defining the functional and organizational information with sufficiently enough abstractions from real chemistry on the structure of the molecules and/or collision rules.

To illustrate intuitively the role of a analytical framework to
analyze emergent phenomena in case of ACs, we consider an AC
consisting of two dimensional polygonal tiles as molecules and the
reaction rule is, if two colliding tiles fit each other on any of
their sides so that the new joined tile has no gap, the resultant
(bigger) tile is included and both of the colliding tiles are
removed, otherwise if they do not fit on any side then they are
discarded. The colliding tiles are chosen at random.

Given this much we can observe and analyze the population of
emerging tiles over a course of time and see if any interesting
organization of molecules emerges or not. We can consider the
possibility of the emergence of self replicating molecules or
organizations. It can be argued that in this AC, self replication
is not possible. This is evident when we analyze the reaction
rules and then learn that the resultant tile is always different
and bigger than the colliding tiles. In this conclusion it is
implicitly assumed that self replication is defined as appearance
of new tile, which is same in shape and size with either of the
tiles participating in the reaction (either in a single step or in
series of steps). But if we change the ``meaning'' of
self-replication as only the replication of shape and do not
consider the size, we can find example of tiles which result in
bigger tile with same shape as one of the colliding tiles. This
example highlights the significance of an analytical framework, in
this case, the analysis of functional information associated with
the tiles as per the reactions rules with respect to meaning of
self replication (context) to conclusively determine whether self
replication will emerge or not. There is more discussion on
self-replication in the section 3.5.

Before we proceed to describe our initial proposal for this
framework, we will review the two important known notions of
information. They are Shannon's Entropy of Information and
Algorithmic Information. This will motivate us to make the point
for the need of a new and broader notion of information.

\subsection{Analysis of the Notions of Information}

Two well known notions of information in computer science are
Shannon's Entropy notion based on coding theory and Algorithmic
Information. Both of these, though, capture certain aspects of
what we call information but not everything.

\subsubsection{Shannon's Information Entropy}

In Shannon's information theory \cite{18} the amount of
information associated with any symbol is the logarithm of the
probability of occurrence of that symbol in a message.

The \emph{Shannon's entropy} of a variable $X$ measured in bits is
defined as

$$H(X) = - \Sigma_{x}P(x)log_2P(x)$$

where $P(x)$ is the probability that $X$ is in the state $x$, and
$Plog_2P$ is defined as $0$ if $P = 0$. The joint entropy of
variables $X_1, \ldots, X_n$ is then defined by

$$H(X_1, \ldots, X_n) = -\Sigma_{x_1} \ldots \Sigma_{x_n} P(x_1,
\ldots, x_n)log_2P(x_1, \ldots, x_n)$$

The mutual information between two discrete random variables X and
Y is defined to be

$$I(X;Y) = H(X) + H(Y) - H(H,Y)$$

bits, where H(X) is the entropy of the random variable X and
H(X,Y) is the joint entropy of these variables.

The Shannon's notion of entropy is quite useful in certain aspects
in molecular biology such as, mutual information as a measure of
the information content of protein families, or statistical
properties of genetic material \cite{11}. Still the basic problem
is that Shannon's entropy has nothing to do with the ``meaning''
of the message as Shannon explained in his original paper in the
beginning itself \cite{18}. This absence of meaning in the Shannon's
notion of information makes a fundamental difference when we
consider for example the genetic material. Precisely because
significance of genetic material is only due to its characteristic
functions inside a cell. These functional properties cannot be
captured by this measure of entropy which is basically used when
information is being transmitted through a channel between a
sender and a receiver. In case of information with evolving
entities even in pre-biotic evolution, what is important is the
formation of information encoding mechanism such as genetic
material and this sender, receiver and channel aspect does not
arise directly. Still, when considering the communication
aspect in our framework, this notion might be of some use.\\

\subsubsection{Algorithmic Information}

Algorithmic information is basically the size of the shortest
program, which can produce the description of the object as binary
string when simulated by a universal computing machine
\cite{15}. This notion of information in algorithmic terms was
motivated to define the notion of randomness precisely. For a
natural number its algorithmic information is equal to its
logarithm. Similarly, for any randomly chosen binary string, its
expected algorithmic information is roughly equal to its length.

Though algorithmic information is quite useful when dealing with
the structural or syntactical aspects of an object's description,
it does not capture the functional properties of the object. For
example, two programs written as binary strings as input for a
universal computer may have nearly the same algorithmic
information but one program may behave be fundamentally different than
the other on execution. Thus algorithmic information says nothing
of the semantic aspect or the context based functional properties
of the object. As another example there might be several molecular
structures of high complexity which might have same algorithmic
information due to their structures as typical genetic material
but may not exhibit same properties when inside a cell as what
genetic material exhibits.

This context based functional properties or the semantics
specified by the context (environment) is something missing from the
current analysis and demands clear formulation. As a remote
analogy we can consider this as somewhat similar to program
equivalence modulo finite execution steps, whereby two programs
executing same way for finitely many steps or structures
exhibiting similar reactions are considered equivalent. We
consider finite time steps because the more general problem of
program equivalence is undecidable and because in real terms it is
only finitely many possibilities where an object is expected to
function. Thus the presence of genetic code amounts to emergence
of an specific kind of structure which can be analyzed correctly
only when we consider the specific context based functional
information associated with it. \cite{bern90} has detailed
discussion on the role of information in the origin of life.

Once properly defined this functional information can be evaluated
from the structure of reaction rules itself, by computing the
change of information from inputs to outputs. In \cite{19, 20} an
information space is considered as a fundamental block for
analysis of evolution of complex form of life in artificial life
systems. They even consider that presence of structure for
programmable information is one of the necessary conditions for
evolution of complexity.

\subsection{Towards  an Abstract Framework}

Based upon the analysis of ACs  and discussion on the relevance of
``context based functional information'', we propose here an
initial sketch for a new framework to study the emergent
phenomenon such as emergence of self replication in molecules,
emergence of hypercycles, metabolic networks, self organization
and other life-like properties from a basic AC set-up in a unified
way.

We identify three basic high level abstractions in our
framework, viz., \emph{information, computation,
 and communication}. These notions need to be further refined and clearly formalizes in the context of ACs and in general AL studies. These are discussed next.

 \begin{description}
    \item[Information]  Among the list of open problems in AL presented
    in \cite{Bedau00} the
    last problem is - \emph{Develop a theory of information processing,
    information flow, and information generation for evolving systems.}
    Information in the framework will be a way to characterize and compare various structures for their relative
     functional or semantic aspects. In living systems, this information has very specific role
     of controlling the way various processes work and transmission of information
     after replication etc. We need to understand if an AC able to create information. Does
``information processing'' emerge by way of evolution?
     Again as noted in \cite{Bedau00} it is important to understand the interaction of
     objects with the environment in terms of information processing and
     information generation.  From the discussion in the previous section
     on two of the known notions of
     information, Shannon's Entropy and Algorithmic Information
     we know that these might not be adequate to capture completely the kind of information
     we are interested in.
    \item[Computation] Extended Chruch - Turing thesis as proposed
    in \cite{exchruchturing} states that all natural processes are basically
    computational in nature or can be reduced in computational terms, although there
    can be fundamental differences in the actual model which
    nature might employ with the known models of computation
    (e.g., Turing machine,
    lambda calculus etc.) This essentially means that we can
     understand the
    problem of origin or life and complex forms in computational terms.
    This computational aspect is
    important because it can be used to understand the generation of
    information
    during the course of evolution under the effect of rule space.
    This way, if we assume that molecules are computational units and their
    interaction gives rise
    to new computational units, we can try to understand how
    the computational dynamics changes and under what conditions
    life-like forms and other relevant aspects emerge. Self- replication
    has already been demonstrated to be a computational process \cite{selfrep}.
     Along the line of computation we can think of the organizational structures
     emerging in ACs as some kind of (distributed) computing
     structures and can analyze changes in their structures in terms computational power.
     Most importantly we expect that it is possibly a powerful computation process which
    gives rise to the emergence of new forms of information processing like genetic material.
    \item[Communication] It is not clear if communication in
the form of explicit signal passing happens in case of molecular
reactions but we may try to capture selective chemical bonding
which is actually determined by geometrical configuration, bond
strengths in a molecule, and other thermodynamic constraints using
communication analogy. This communication could in turn be used by chemical
metaphors for preferential interaction not explicitly defined by
interaction rules. The communication pathways can be taken into
consideration as a matter of efficiency of evolving of the
organization and especially for distributed and collective
information processing.

 \end{description}

\subsection{Discussion}

Because of the differences in the basic formulation of various ACs
resulting in quite different structures and nature of emergent
phenomena, it is not easy to directly formulate the unified
framework precisely. Therefore we adopt two broad strategies to
achieve the goal, which can be termed as bottom up approach and
top down approach . \\

\emph{Inductive Approach}. This is the approach to formulate first
the framework for specific ACs and then combine these individual
 formulations into more general formulation of the unified model.

 To start with, a P system based Artificial Graph Chemistry, discussed in~\cite{agc}, could be used
 initially for the formulation of this framework and to do experiments
 to refine and verify that further. Hence the first step is to carry out
experiments with the P system based AC. If life-like structures
are observed to be emerging, study the conditions leading to their
emergence from the point of unified framework. This step will
elaborate the possible principles of emergence of life and will
assist in possible refinement and formulation of the framework.

In next step one could work towards generalizing this formulation
with respect to essential features in major ACs. This can be done
either by attempting to generalize this formulation and then
verify it for other ACs or by formulating separately for other ACs
and then combining them together to come to a more general
formulation. Further experiments could be carried out to refine the
formulation. In this approach the main problem is to select right
ACs and then to understand them clearly to come to a formulation. \\

\emph{Deducutive Approach}. In this approach we aim to come to a
generic formulation first in analytical way by analyzing the
essential features of life-like emergent phenomena.

The computation theoretic framework based on cellular automata for
self replication is one such example of this kind of top down
approach \cite{selfrep}. The quantitative analysis of ACs with
self replicating structures is usually done using generic
\emph{replicator-equations} of population dynamics \cite{5}. A
replicator-equations for a population consisting of $n$ species
with relative frequencies $x_1, x_2, \ldots, x_n$ is formally
given by,

$$\dot{x_i} = x_i(f_i(\mathbf{x}) - \sum_{j=1}^{n}x_jf_j(\mathbf{x}))$$

where for each species $i$, $f_i$ describes the \emph{fitness} of
$i$. These fitness functions are usually taken to be linear.
$\mathbf{x}$ is a vector $(x_1, x_2, \ldots, x_n)$.

As an another example of top down approach, we consider
hypercycles, which are roughly the set of small self-replicating
molecules where reaction of each molecule feeds the production of
some other molecule in a cyclic fashion. This cyclic dependence of
the self replicating molecules gives rise to bigger self
replicating structures. The hyper cycles were introduced and
formally characterized using reactor flow equations in
\cite{hypercycles}. That characterization is general enough to
capture any kind of population dynamics. Though again this is
quantitative characterization and cannot be used to explain why
hyper cycles actually emerge or whether they will emerge at all in
an organization where new species keep emerging.

To take this approach further, we identify the following basic
elements in emergence of self-replication in an AC set-up.

\begin{description}
    \item [Identity] - these are the most elementary entities
    of replication, that is, which self-replicate itself. Examples of individual
    cells in an multi-cellular organism are such examples. In real
    chemistry we notice that, though atoms are the basic
    components (of self-replicating entities), they do not self-replicate. Thus
    identification of these self-replicating entities is important
    to understand any level of self-organization. This is not easy
    always because there is no bound on the "size" or "type" of
    these replicating molecules. This might be the case that there
    are several hierarchies of self- replicating entities, each
    replicating on its level.

    \item  [Self-preservation] - this means structure is robust against
    perturbations and thus small changes in the structure cannot
    be taken for dissimilarity. Before talking about replication,
    the entities need to be able to preserve their own identity.
    How do we assign an identity to the entities that is preserved over time?

     \item [Causality]  Let us say a new entity appears somehow.
     How do we establish that the new entity arises from an existing entity?
     Only if we establish a causal relationship between the old
    and new entity, we would be able to talk about self-replication.

    \item [Equivalence Relation] - This relation is used to
    correctly formulate the characteristics, which will be used to
    determine the presence of replication. To clarify the point,
    again consider the case of replicating cells, there not
    everything replicates itself during cell division, therefore similarity
    in overall chemical composition or equal cell sizes cannot be the
    basis of characterizing self-replication. In fact it is mainly genetic material which
    replicates during cell division and we treat is as cell replication.

    A working equivalence relation can be functional equivalence using some
    measure of ``context based functional information'', that is, if two
    structures can function the same way then they are treated
    equivalent. The problem, as we discussed in last section correct
    formulation of this information.

    \item [Period of replication] - this is measured to find out
    after how many reaction steps, a self-replicating structure
    will replicate itself. In most of the simple cases it is just one
    reaction period, which means structures maintain and replicate themselves
    for each reaction. It need not to be the case for a larger
    self-replicating organization, which might involve
    gradual replication of its components across several reaction
    cycles.

\end{description}

As second step we will verify and refine this generic formulation
based upon the experiments with individual ACs, for example to see
whether this formulation can be used for analyzing the emerging
structures in P system based AC.

\section{Final Discussion}

Objective of this section is to summarize main goals of the
proposal and discuss the broader picture where these goals may fit
in an AC research.

To summarize, we aim to develop a unified formal framework to
    characterize and classify symbolic ACs based upon three high
level abstractions viz., information, computation and
communication. This framework will be used to understand the
aspect of information processing and computation in an AC
environment.

There are two major approaches to develop the unified framework.
The bottom-up or inductive approach is by formulating analytical framework for
individual ACs and then combining them to more general one. Second
top-down or deductive approach is to work for general formulation with later
refinement and verification using actual AC set-ups.

ACs are basically designed to complement the main stream AL
research. This is primarily because major AL studies presume the
prior existence of basic structure of life-like entities and
develop over them. This leaves the question of origin of these
basic structures open and that is where ACs come into picture.

Because the main theme of AL research is to discover the possible
biological principles which might be working independent of
physical laws, AL studies mainly draw motivation from real-life
biological phenomena. Theory of evolution based on random
mutations and fitness based natural selection is one such source
of motivation in many AL studies \cite{1}. Similarly ACs also draw
motivation from real chemistry. \emph{The main conceptual
motivation ACs borrow from real chemistries is not the actual
chemical structures or reactions but the abstract concept that
life originated as a result of complex dynamical interplay between the rule
space consisting of reaction rules or semantics and the object space consisting
of the molecules which react.} This is what is the prime source of
differences in various ACs in their definition and structures,
since there is no such generic framework which can used by ACs to
define the basic structure of molecules or reactions. Most often
what is clear is only the end results, that is, an AC set up is
expected to lead to the emergence of certain basic characteristics
of life, e.g., self-replication, metabolism etc.


\end{document}